\documentclass{article}
\usepackage{graphicx} 
\usepackage{multirow}
\usepackage{parskip}
\usepackage{caption}
\usepackage{hyperref}
\hypersetup{hidelinks}
\usepackage[font={small}, labelfont={bf}]{caption}
\usepackage[backend=bibtex,style=nature]{biblatex}

\usepackage[margin=1in]{geometry}
\usepackage{tikz}
\usetikzlibrary{shapes.multipart, arrows.meta}
\usetikzlibrary{positioning}
\title{The Perils \& Promises of Fact-checking with Large Language Models}

\author{
    Dorian Quelle \& Alexandre Bovet \\[0.5cm]
    Department of Mathematical Modeling and Machine Learning, \\
    Digital Society Initiative,\\
    University of Zurich, Zurich, Switzerland\\[0.5cm]
    dorian.quelle@uzh.ch, 
    alexandre.bovet@uzh.ch
}

\date{October 2023}

\begin{document}

\maketitle

\section*{Abstract}
Automated fact-checking, using machine learning to verify claims, has grown vital as misinformation spreads beyond human fact-checking capacity. Large Language Models (LLMs) like GPT-4 are increasingly trusted to write academic papers, lawsuits, and news articles and to verify information, emphasizing their role in discerning truth from falsehood and the importance of being able to verify their outputs.
Understanding the capacities and limitations of LLMs in fact-checking tasks is therefore essential for ensuring the health of our information ecosystem.
Here, we evaluate the use of LLM agents in fact-checking by having them phrase queries, retrieve contextual data, and make decisions.
Importantly, in our framework, agents explain their reasoning and cite the relevant sources from the retrieved context.
Our results show the enhanced prowess of LLMs when equipped with contextual information. GPT-4 outperforms GPT-3, but accuracy varies based on query language and claim veracity. While LLMs show promise in fact-checking, caution is essential due to inconsistent accuracy. Our investigation calls for further research, fostering a deeper comprehension of when agents succeed and when they fail.

\section{Introduction}
Fact-checking has become a vital tool to reduce the spread of misinformation online, shown to potentially reduce an individual's belief in false news and rumors \cite{porter2021global,morris2020fake} and to improve political knowledge \cite{nyhan2015estimating}.
While verifying or refuting a claim is a core task of any journalist, a variety of dedicated fact-checking organizations have formed to correct misconceptions, rumors, and fake news online. 
A pivotal moment in the rise of fact-checking happened in 2009 when the prestigious Pulitzer Prize in the national reporting category was awarded to Politifact. Politifact's innovation was to propose the now standard model of an ordinal rating, which added a layer of structure and clarity to the fact check and inspired dozens of projects around the world \cite{mantzarlis2018fact}. 
The second wave of fact-checking organizations and innovation in the fact-checking industry was catalyzed by the proliferation of viral hoaxes and fake news during the 2016 US presidential election \cite{Bovet2019,Grinberg2019} and Brexit referendum \cite{mantzarlis2018fact}. Increased polarization \cite{flamino_political_2023}, political populism, and awareness of the potentially detrimental effects of misinformation have ushered in the ``rise of fact-checking" \cite{graves2016rise}. 

Although fact-checking organizations play a crucial role in the fight against misinformation, notably during the COVID-19 pandemic \cite{siwakoti2021covid},
the process of fact-checking a claim is an extremely time-consuming task. A professional fact-checker might take several hours or days on any given claim \cite{hassan2015quest,adair2017progress}. Due to an ever-increasing amount of information online and the speed at which it spreads, relying solely on manual fact-checking is insufficient and makes automated solutions and tools that increase the efficiency of fact-checkers necessary.

Recent research has explored the potential of using large artificial intelligence language models as a tool for fact-checking \cite{hoes2023leveraging, sawinski2023openfact, he2021debertav3,caramancion2023news,choi2023automated}. However, significant challenges remain when employing large language models (LLMs) to assess the veracity of a statement. One primary issue is that fact-checks are potentially included in some of the training data for LLMs. Therefore, successful fact-checking without additional context may not necessarily be attributed to the model's comprehension of facts or argumentation. Instead, it may simply reflect the LLM's retention of training examples. While this might suffice for fact-checking past claims, it may not generalize well beyond the training data.

Large language models (LLMs) like GPT-4 are increasingly trusted to write academic papers, lawsuits, news articles\footnote{\url{https://cybernews.com/news/academic-cheating-chatgpt-openai/},\\\url{https://www.nytimes.com/2023/06/08/nyregion/lawyer-chatgpt-sanctions.html}}, or to gather information \cite{choudhury2023investigating}. Therefore, an investigation into the models' ability to determine whether a statement is true or false is necessary to understand whether LLMs can be relied upon in situations where accuracy and credibility are paramount. The widespread adoption and reliance on LLMs pose both opportunities and challenges. As they take on more significant roles in decision-making processes, research, journalism, and legal domains, it becomes crucial to understand their strengths and limitations. The increasing use of advanced language models in disseminating misinformation online highlights the importance of developing efficient automated systems. The 2024 WEF Global Risk Report ranks misinformation and disinformation as the most dangerous short-term global risk as LLMs have enabled an “explosion in falsified information” removing the necessity of niche skills to create “synthetic content” \cite{WEF2024}.  On the other hand, artificial intelligence models can help identify and mitigate false information, thereby helping to maintain a more reliable and accurate information environment. The ability of LLMs to discern truth from falsehood is not just a measure of their technical competence but also has broader implications for our information ecosystem. 

A significant challenge in automated fact-checking systems relying on machine learning models has been the lack of explainability of the models' prediction. This is a particularly desirable goal in the area of fact-checking as explanations of verdicts are an integral part of the journalistic process when performing manual fact-checking \cite{kotonya2020explainable}. While there has been some progress in highlighting features that justify a verdict, a relatively small number of automated fact-checking systems have an explainability component \cite{kotonya2020explainable}. 

Since the early 2010s, a diverse group of researchers have tackled automated fact-checking with various approaches. This section introduces the concept of automated fact-checking and the different existing approaches. Different shared tasks, where research groups tackle the same problem or dataset with a defined outcome metric, have been announced with the aim of automatically fact-checking claims. For example, the shared task \textsc{RumourEval} provided a dataset of ``dubious posts and ensuing conversations in social media, annotated both for stance and veracity'' \cite{rumoreval}. \textsc{CLEF CheckThat!} prepared three different tasks, aiming to solve different problems in the fact-checking pipeline \cite{nakov2022overview}. First, ``Task 1 asked to predict which posts in a Twitter stream are worth fact-checking, focusing on COVID-19 and politics in six languages'' \cite{cleftask1}. Task 2 ``asks to detect previously fact-checked claims (in two languages)'' \cite{cleftask2}. Lastly, ``Task 3 is designed as a multi-class classification problem and focuses on the veracity of German and English news articles'' \cite{cleftask3}. The Fact Extraction and VERification shared task (\textsc{FEVER}) ``challenged participants to classify whether human-written factoid claims could be SUPPORTED or REFUTED using evidence retrieved from Wikipedia'' \cite{fever}. In general, most of these challenges and proposed solutions disaggregate the fact-checking pipeline into a multi-step problem, as detection, contextualization, and verification all require specific approaches and methods \cite{das2023state}. For example, \cite{hassan2017claimbuster} proposed four components to verify a web document in their \textsc{ClaimBuster} pipeline. First, a claim monitor that performs document retrieval (1), a claim spotter that performs claim detection (2), a claim matcher that matches a detected claim to fact-checked claims (3), and a claim checker that performs evidence extraction and claim validation (4) \cite{zeng2021automated}.

In their summary of automated fact-checking \cite{zeng2021automated} define entailment as ``cases where the truth of hypothesis $h$ is highly plausible given text $t$''. More stringent definitions that demand that a hypothesis is true in ``every possible circumstance where $t$ is true'' fail to handle the uncertainty of Natural Language. Claim verification today mostly relies on fine-tuning a large pre-trained language model on the target dataset \cite{zeng2021automated}. State-of-the-art entailment models have generally relied on transformer architecture such as BERT \cite{devlin2018bert} and RoBERTa \cite{liu2019roberta}. \cite{hoes2023leveraging} tested GPT-3.5's claim verification performance on a dataset of PolitiFact statements without adding any context. They found that GPT-3.5 performs well on the dataset and argue that it shows the potential of leveraging GPT-3.5 and other LLMs for enhancing the efficiency and expediency of the fact-checking process. Novel large language models have been used by \cite{sawinski2023openfact} in assessing the check-worthiness. The authors test various models ability to predict the check-worthiness of English language content. The authors compared GPT-3.5 with various other language models. They find that a fine-tuned version of GPT3.5 slightly ourperforms DeBerta-v3 \cite{he2021debertav3}, an improvement over the original DeBERTa architecture \cite{he2020deberta}. \cite{choi2023automated} use fact-checks to construct a synthetic dataset of contradicting, entailing or neutral claims. They create the synthetic data using GPT-4 and predict the entailment using a smaller fine-tuned LLM. Similarly, \cite{caramancion2023news} test the ability of various LLMs to discern fake news by providing Bard, BingAI, GPT-3.5 and GPT-4 on a list of 100 fact-checked news items. The authors find that all LLMs achieve performances of around 64-71\% accuracy, with GPT 4 receiving the highest score among all LLMs. \cite{cuartielles2023retraining} interview fact-checking platforms about their expectations of Chat-GPT as a tool for both misinformation fabrication, detection, and verification. They find that while professional fact-checkers highlight the potential perils such as the reliability of sources, the lack of insights into the training process, and the enhanced ability of malevolent actors to fabricate false content, they nevertheless view it as a useful resource for both information gathering and the detection and debunking of false news \cite{cuartielles2023retraining}.

While earlier efforts in claim verification did not retrieve any evidence beyond the claim itself (for example, see \cite{rashkin2017truth}), augmenting claim verification models with evidence retrieval has become standard for state-of-the-art models \cite{Guo2021ASO}. In general, evidence retrieval aims to incorporate relevant information beyond the claim. For example, from encyclopedias (e.g., Wikipedia \cite{thorne2018fever}), scientific papers \cite{wadden2020fact}, or search engines such as Google \cite{augenstein2019multifc}. \cite{augenstein2019multifc} submit a claim verbatim as a query to the Google Search API and use the first ten search results as evidence. A crucial issue for evidence retrieval lies in the fact that it implicitly assumes that all available information is trustworthy and that veracity can be gleaned from simply testing the coherence of the claim with the information retrieved. An alternative approach that circumvents the issue of the inclusion of false information has been to leverage knowledge databases (also knowledge graphs) that aim to ``equip machines with comprehensive knowledge of the world's entities and their relationships'' \cite{Weikum2020MachineKC}. However, this approach assumes that all facts pertinent to the checked claim are present in a graph. An assumption that \cite{Guo2021ASO} called unrealistic. 

Our primary contributions in this study are two-fold. First, we conduct a novel evaluation of two of the most used LLMs, GPT-3.5 and GPT-4, on their ability to perform fact-checking using a specialized dataset. An original part of our examination distinguishes the models' performance with and without access to external context, highlighting the importance of contextual data in the verification process. Second, by allowing the LLM agent to perform web searches, we propose an original methodology integrating information retrieval and claim verification for automated fact-checking. By leveraging the ReAct framework, we design an iterative agent that decides whether to conclude a web search or continue with more queries, striking a balance between accuracy and efficiency. This enables the model to justify its reasoning and cite the relevant retrieved data, therefore addressing the verifiability and explainability of the model's verdict. Lastly, we perform the first assessment of GPT-3.5's capability to fact-check across multiple languages, which is crucial in today's globalized information ecosystem. 

We find that incorporating contextual information significantly improves accuracy. This highlights the importance of gathering external evidence during automated verification. We find that the models show good average accuracy, but they struggle with ambiguous verdicts. Our evaluation shows that GPT-4 significantly outperforms GPT-3.5 at fact-checking claims. However, performance varies substantially across languages. Non-English claims see a large boost when translated to English before being fed to the models. We find no sudden decrease in accuracy after the official training cutoff dates for GPT-3.5 and GPT-4. This suggests that the continued learning from human feedback may expand these models' knowledge.

\section{Materials and methods}

\subsection{Approach}
This paper contributes to both the evidence retrieval and claim verification steps of the automated fact-checking pipeline. Our approach focuses on verifying claims -- assessing if a statement is true or false, while simultaneously retrieving contextual information to augment the ability of the LLM to reason about the given claims. We use large artificial intelligence language models GPT-3.5 and GPT-4. We combine state-of-the-art language models, iterative searching, and agent-based reasoning to advance automated claim verification.

GPT-3.5 and GPT-4 are neural networks trained on vast amounts of textual data to generate coherent continuations of text prompts \cite{vaswani2017attention}. They comprise multiple layers of transformer blocks, which contain self-attention mechanisms that allow the model to learn contextual representations of words and sentences \cite{vaswani2017attention}. LLMs are trained using self-supervision, where the model's objective is to predict the next token in a sequence of text. The GPT models are trained on vast amounts of unstructured textual data like the common crawl dataset, which is the largest dataset to be included in the training. Common Crawl is a web archive that consists of terabytes of data collected since 2008 \cite{buck2014n} \cite{brown2020language}. GPT-3.5 and GPT-4 are additionally trained with reinforcement learning from human feedback (Rlhf), where human feedback is incorporated to enhance the models' usability. OpenAI states that they regularly update their models based on human feedback, potentially leading to knowledge of current events that expands upon the initial training regime, which was stopped in September of 2021\footnote{\url{https://openai.com/blog/chatgpt}}.  

\begin{figure}[htbp]
\begin{center}
\includegraphics[width=15cm, trim={1cm 3.5cm 1cm 3.5cm}, clip]{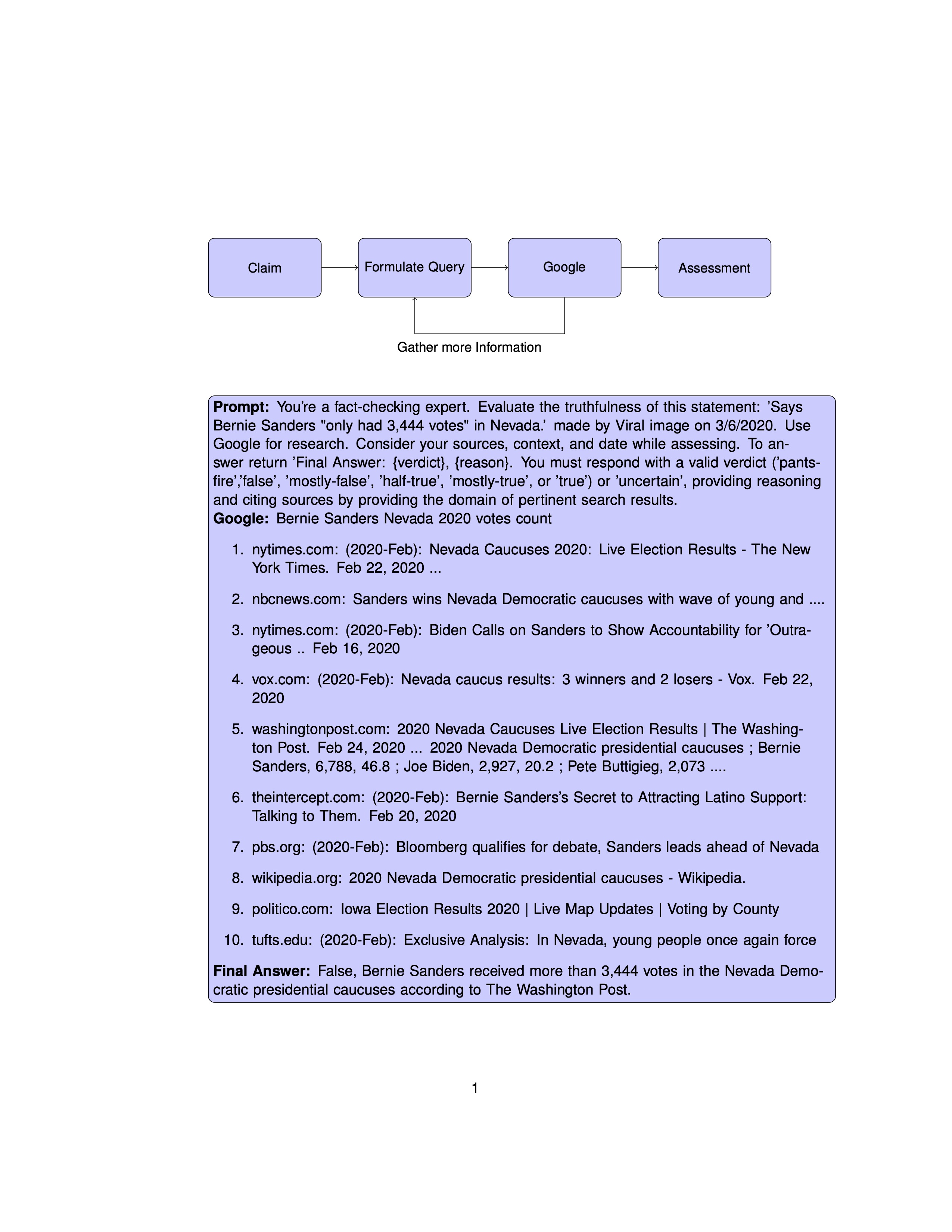}
\end{center}
\caption{Workflow showing how we enable LLM agents to interact with a context to assess the veracity of a claim (top). Example of the treatment of a specific claim (bottom)}\label{fig:1_workflow}
\end{figure}

We are evaluating the performance of GPT-3.5 and GPT-4 based on two conditions. First, we query the models with the statement, the author, and the date of the statement. The model does not possess any means of retrieving further information that might enable it to make an informed decision. Rather this approach relies on the model having knowledge of the events described in the claim. In the second condition, we enable the LLM to query a Google Search engine to retrieve relevant information surrounding the claim. To prevent the model from uncovering the fact-check itself, we filter the returned Google search results for all domains present in the dataset. We present the LLM with information returned from the Google Search Engine API, comprising previews of search results. These previews included the title of the website, the link, and an extract of relevant context, mirroring the typical user experience on Google. To refine our approach, we experimented with integrating additional information from the full HTML content of websites. We quickly realised that this voluminous data was overwhelming the LLM's context window, leading to suboptimal performance. To address this, we employed BM25, an information retrieval function \cite{robertson2009probabilistic}, to distil the most critical parts of each website. We found no improvement to the performance of the LLMs, as Google already uses machine learning methods to identify the most important information to include in the preview. We equip the agent with the ability to query Google by leveraging the Reasoning \& Acting (\emph{ReAct}) framework, proposed by \cite{yao2023react}, which allows an LLM to interact with tools. The goal of the ReAct framework is to combine reasoning and take actions. Reasoning refers to the model planning and executing actions based on observations from the environment. Actions are function calls or API calls by the model to retrieve additional information from external sources. The model is initially prompted with the claim and then decides if it needs to take actions. As we equip both models with the ability to query Google, the model then is able to retrieve information and receives its next observation. Based on this observation, the model can decide to either return a final answer or retrieve information with a different query. If the model fails to answer after three iterations of retrieving information, we terminate the search. The model retrieves 10 Google search results per iteration. To test the claim verification in a realistic scenario, any result from a fact-checking website present in the dataset is removed from the results. If the model was terminated because it reached the maximum number of iterations, it is prompted to provide a final answer based on all of its previous Google searches. In this paper, we employ the LangChain library \cite{Chase_LangChain_2022} to create an agent within the ReAct framework. We show in Fig. \ref{fig:1_workflow} the workflow we implemented and an example of the treatment of a claim.

We employed the 16k context window which is standard for the GPT API. Since our experiments larger context windows have been introduced  \cite{OpenAIModels}, which could enable future LLM powered fact-checking applications to incorporate more search engine results or to include more content from each website.

To illustrate the capabilities of the proposed system, Fig. \ref{ExampleCorrect} and Fig. \ref{ExampleIncorrect} showcase two correctly classified and two incorrectly assessed verdicts. The full transcripts from these examples, including all retrieved Google links, are available in the supplementary material.

\setcounter{figure}{1}
\begin{figure}[htbp]
\begin{center}
    \includegraphics[width=0.75\textwidth]{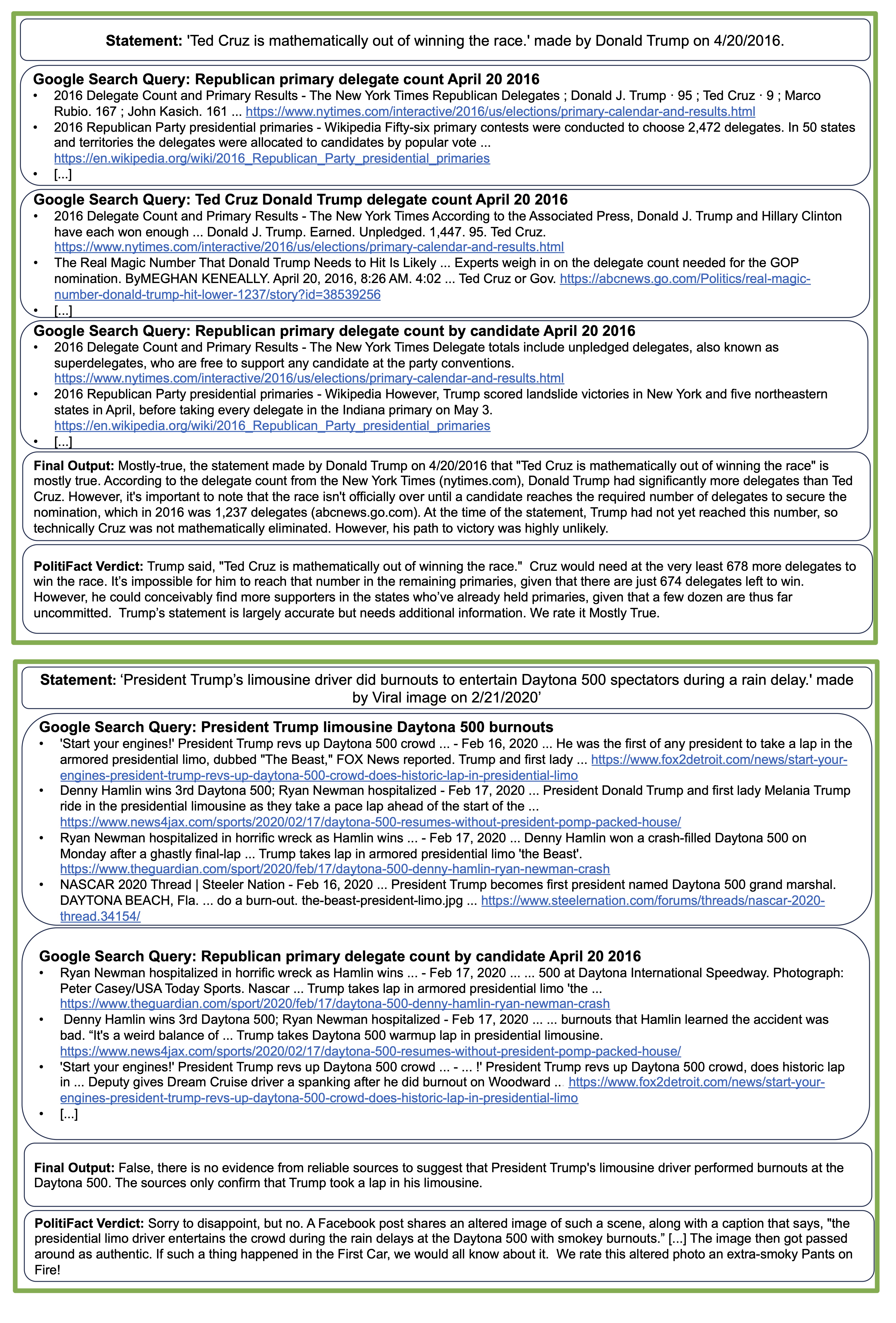}
\end{center}
\caption{Examples of Politifact statements for which the model returns correct responses. The LLM is tasked with verifying a statement made by Donald Trump, indicating that Ted Cruz is mathematically out of the race. The LLM uses Google to retrieve information on the delegate count and correctly concludes the statement is mostly true. We show the Google queries performed by the LLM and the first results of each query. In the second example, the LLM is tasked to verify a statement claiming Donald Trump's driver "did burnouts" during a race. The LLM finds information that Donald Trump did a lap around the race but correctly concludes that no information indicates that he did "burnouts". The full examples, including all Google results, are shown in the Supplementary Material.}
\label{ExampleCorrect}
\end{figure}

\setcounter{figure}{2}
\begin{figure}[htbp]
\begin{center}
    \includegraphics[width=0.75\textwidth]{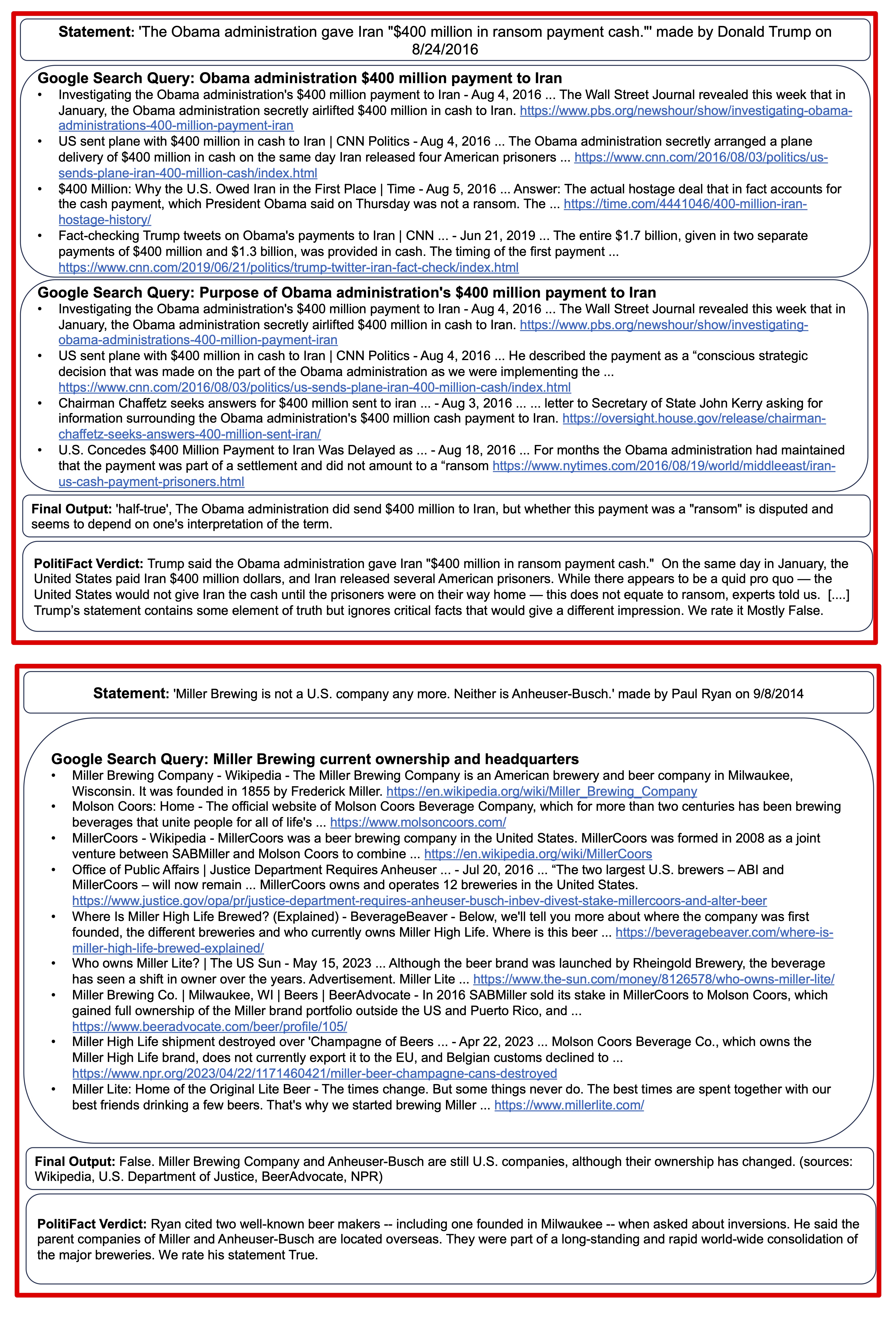}
\end{center}
\caption{Examples of Politifact statements for which the model returns incorrect responses. We show the Google queries performed by the LLM and the first results of each query. The LLM is asked to verify whether the Obama administration paid a ransom payment to Iran. The LLM finds information on the payment but can't conclusively confirm the purpose of the payment. It concludes that the statement is half-true. PolitiFact argues that the statement is mostly-false, as the payment is not necessarily a ransom payment. In the second example, the LLM is asked to verify whether a beer brand is American.  It finds information indicating that the company is American and returns False. The company has, however, been bought by foreign investors, making the statement true. The full examples, including all Google results, are shown in the Supplementary Material.}
\label{ExampleIncorrect}
\end{figure}

\subsection{Experiment}
We conduct two experiments to evaluate how well our agents can fact-check different claims. First, we evaluate whether GPT-4 is significantly better at fact-checking than GPT-3.5. GPT-4 has been shown to outperform GPT-3.5 on a variety of benchmarks, particularly in zero-shot reasoning tasks \cite{espejel2023gpt}. We therefore compare the performance of GPT-3.5 and GPT-4 on two datasets, a dataset of US political fact-checked claims provided by PolitiFact \cite{misra2022politifact} and a dataset of multilingual fact-checked claims provided by Data Common. As GPT-3.5 is openly available to the public for free, it is used substantially more used than GPT-4, thereby increasing the importance of understanding its performance.

\subsubsection{Experiment on the PolitiFact Dataset}
The dataset used in this experiment consists of a database of fact-checked claims by PolitiFact \cite{misra2022politifact}. Each observation has a statement originator (the person who made the statement being fact-checked), a statement (the statement being fact-checked), a statement date, and a verdict. The verdict of a fact-check is one of six ordinal categories, indicating to which degree a statement is true or false: True, Mostly-True, Half-True, Mostly-False, False, and Pants-Fire. Pants-Fire indicates that a statement is utterly false. In total, the dataset has 21,152 fact-checks, spanning a time-frame of 2007 to 2022. The number of fact-checks per month is shown in Fig. \ref{FactsPerMonth}. We sampled 500 claims for each response category, leading to a total of 3000 unique fact-checks to be parsed. We thereby ensured an equal distribution of outcome labels in the final dataset. 

While comparing the two models, we also compare the accuracy of the model with and without additional context. We therefore run four conditions, GPT-3.5 with context and without context, and GPT-4 with and without context. The \emph{No-Context} condition refers to the model being prompted to categorize a claim into the veracity categories, without being given any context, or the ability to retrieve context. In addition to the veracity labels, the model is able to return a verdict of "uncertain" to indicate, that it isn't capable of returning an assessment. In the \emph{Context} condition, the agent is capable of formulating Google queries to retrieve information pertinent to the claim. When the model is able to retrieve contextual information, any results from PolitiFact are excluded from the results.

\subsubsection{Experiment on the Multilingual Dataset}
Secondly, we investigate whether the ability of LLMs to fact-check claims depends significantly on the language of the initial claim. While both GPT-3.5 and GPT-4 exhibit impressive multilingual understanding, a variety of empirical results have shown that large language models struggle to adequately understand and generate non-English language text \cite{bang2023multitask}, \cite{jiao2023chatgpt}, \cite{zhu2023multilingual}. However, misinformation is a problem that is not restricted to high-resource languages. Conversely, fact-checking has become a global endeavor over the last decade, with a growing number of dedicated organizations in non-English language countries.

The dataset used is a fact-check dump by Data Commons. Each fact-check has an associated author and date. Additionally, it contains the \emph{Claim} that is being fact-checked, a \emph{Review}, and a \emph{verdict}. 

The dataset contains a breadth of fact-checking organizations and languages. In total, we detect 78 unique languages in the dataset. By extracting the domain of each fact-check link we find that it contains fact-checks from 454 unique fact-checking organizations. The largest domains, factly (7277), factcrescendo (5664), youturn (3582), boatos (2829), dpa-fact-checking (2394), verafiles (1972), uol (1729), and tempo (1133), have more than one thousand fact-checks, and make up 73\% of the dataset. In  contrast, there are 266 different domains which only have one associated fact-check. Figure \ref{FactsPerMonth} shows the total amount of fact-checks in the dataset per month. While the oldest fact-check in the dataset was published in 2011, we see that very few fact-checks were uploaded until early 2019. Since then the dataset contains a relatively stable amount of around eight hundred fact-checks per month. 

To use the Data Commons dataset we need to standardize the dataset. We reduced the number of unique verdicts, as many different fact-checking organizations use different ordinal scales that are hard to unify. By mapping all scales onto a coarse 4-level scale ("False", "Mostly False", "Mostly True", \& "True"), we can then compare how the ability of our model to fact-check depends on the correct verdict assigned by the fact-checker. We translated all present verdicts from their original language to English. We then manually mapped all verdicts which appeared at least twice in the dataset (n = 468) to the four categories.  We removed all observations that we were unable to map to one of these four categories. This included observations with a verdict such as "Sarcasm", "Satire", or "unconfirmed". 

Subsequently, we discarded all languages, that did not have at least 50 observations and languages that did not have at least 10 "true" or "mostly true" observations. From the remaining languages, we sampled up to 500 observations. To compare the ability of our approach across languages, we then used Googletrans\footnote{\url{https://pypi.org/project/googletrans/}}, a free python library that utilizes the Google Translate API, to translate all claims from their original language to English.

In this experiment, we compare the performance of GPT-3.5 both with context and without context in English and in the original language. In the context condition, we removed all fact-checking websites from the dataset from the Google search results.

\setcounter{figure}{3}
\begin{figure}[h]
\includegraphics[width=\textwidth]{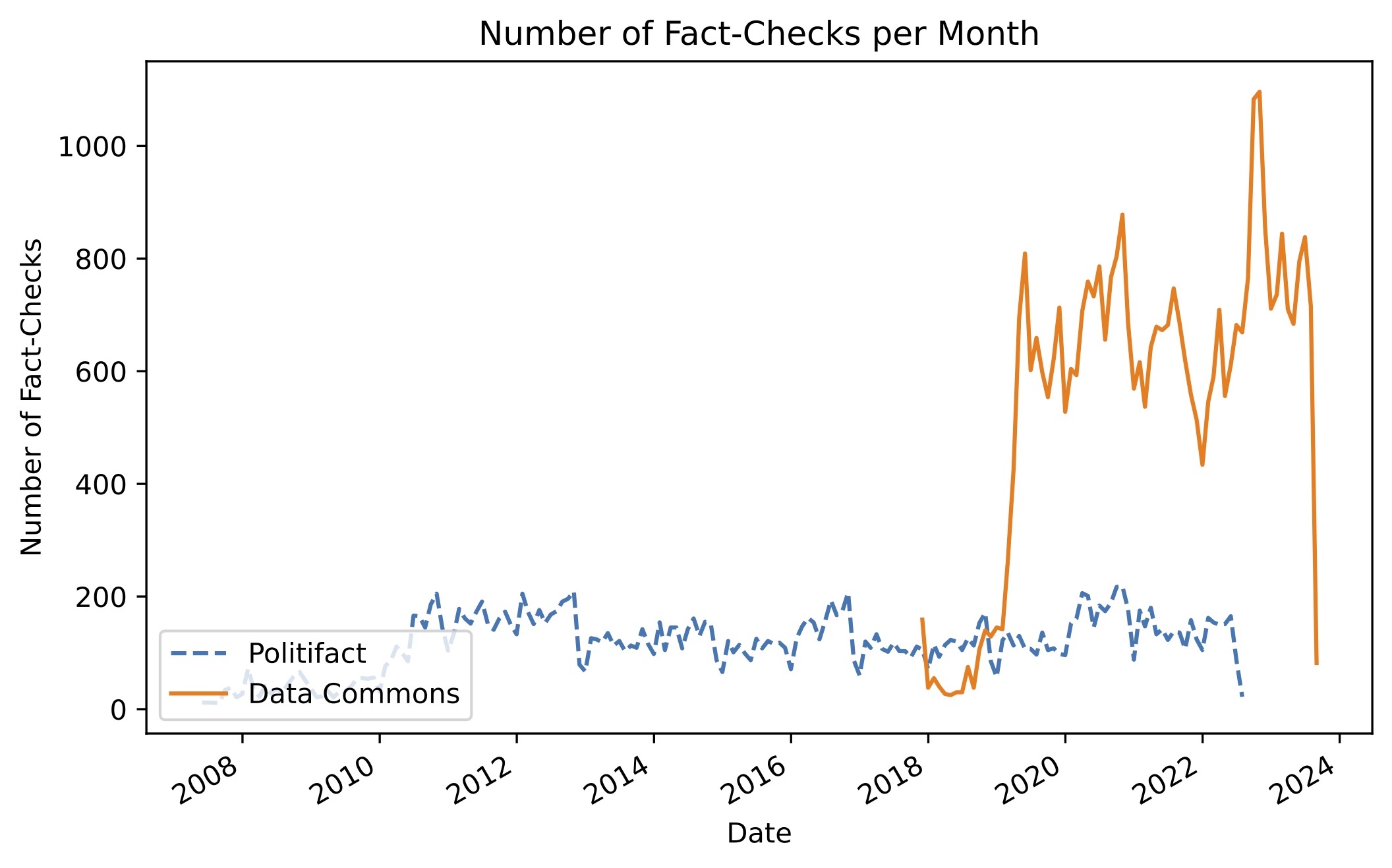}
\caption{Number of fact-checks per month in the Data Commons \& PolitiFact datasets. Number of fact-checks Per Month in the Data Commons \& PolitiFact Datasets. In blue (dashed) the number of fact-checks in the PolitFact Dataset are shown. The orange (solid) line indicates the number of fact-checks in the Data Commons dataset.}
\label{FactsPerMonth}
\end{figure}

\section{Results}

\subsection{Experiment on the PolitiFact Dataset}
\begin{table}[!htbp]
    \centering
    \footnotesize
    \begin{minipage}{0.4\textwidth}
        \centering
            \begin{tabular}{@{\extracolsep{5pt}} cccccccc} 
            \\[-1.8ex]\hline 
            \hline \\[-1.8ex] 
             & \multicolumn{2}{c}{No Context} & & \multicolumn{2}{c}{Context} \\ 
            \cline{2-3} \cline{5-6}\\[-1.8ex] 
            Correct Verdict & GPT-3.5 & GPT-4 & & GPT-3.5 & GPT-4 \\ 
            \hline \\[-1.8ex] 
            pants fire & $92.19$ & $91.80$ & & $93.42$ & $93.92$ \\ 
            false & $81.63$ & $88.08$ & & $88.75$ & $86.13$ \\ 
            mostly false & $79.64$ & $68.70$ & & $71.11$ & $55.82$ \\ 
            half true & $36.16$ & $51.90$ & & $51.58$ & $67.26$ \\ 
            mostly true & $42.35$ & $79.29$ & & $67.27$ & $80.88$ \\ 
            true & $49.11$ & $71.30$ & & $67.61$ & $84.92$ \\ 
            \hline \\[-1.8ex] 
            \end{tabular} 
    \end{minipage}
    \hfill
    \begin{minipage}{0.4\textwidth}
        \centering
            \begin{tabular}{@{\extracolsep{5pt}} ccccccc} 
            \\[-1.8ex]\hline 
            \hline \\[-1.8ex] 
             \multicolumn{2}{c}{No Context} & & \multicolumn{2}{c}{Context} \\ 
            \cline{1-2} \cline{4-5}\\[-1.8ex] 
            GPT-3.5 & GPT-4 & & GPT-3.5 & GPT-4 \\ 
            \hline \\[-1.8ex] 
            $25.81$ & $28.94$ & & $0.00$ & $12.17$ \\ 
            $64.17$ & $47.49$ & &  $86.32$ & $55.82$ \\ 
            $8.82$ & $50.00$ & &  $10.67$ & $40.75$ \\ 
            $2.75$ & $11.58$ & &  $0.00$ & $3.57$ \\ 
            $9.41$ & $59.91$ & &  $36.82$ & $61.13$ \\ 
            $37.05$ & $22.22$ & &  $33.80$ & $33.00$ \\ 
            \hline \\[-1.8ex] 
    \end{tabular} 
    \end{minipage}

    \caption{Comparison of accuracy of all conditions on the PolitiFact dataset. For the table on the left, accuracy is calculated as the percentage of time predicting the correct overall category. Predicting any of pants-fire, false, or mostly false is correct when the claim is labeled as pants-fire, false, or mostly false. Conversely, predicting any half-true, mostly true, or true is correct when the claim is labeled as one of these categories. For the table on the right, accuracy is calculated as the percentage of time predicting the PolitiFact category. For example, predicting "half-true" when the claim is labeled as "mostly-true" is treated as a false prediction.}
    \label{AccGPTPolitifact}
\end{table}

Table \ref{AccGPTPolitifact} presents a comparative analysis of the GPT-3.5 and GPT-4 models in the task of automated fact-checking using the PolitiFact dataset. The table categorizes the performance of both models across different veracity labels ranging from "pants-fire" to "true." In the left table, the accuracy of the model is computed by considering only two categories, i.e. grouping "pants-fire", "false", and "mostly-false" together to false and "half-true", "mostly-true", and "true" to true. In the right table, the accuracy is computed using all the PolitiFact categories.

In the no-context condition, considering the overall accuracy (left table), GPT-4 generally outperforms GPT-3.5. GPT-3.5 predicts the veracity of a claim to be false in 58.2\% of the cases (compared to 22.89\% for GPT-4) achieving an accuracy of 36-49\% in the true-label categories. In the false-label categories, the difference between the two models is significantly smaller and GPT-3.5 outperforms GPT-4 in the mostly false category. Both models achieve an accuracy of over 90\% for claims that are labeled with "pants-on-fire". 
In the context condition, GPT-3.5 is significantly better calibrated, meaning it exhibits a more balanced accuracy between true and false verdicts. While the accuracy in the false categories only differs insignificantly, GPT-3.5 achieves significantly better results in predicting true verdicts in the context condition than in the no-context condition. Similarly, the difference in false categories is smaller for GPT-4 than in true categories, where it achieves an increase of 10.19 percentage points on average. 
In the context condition, both GPT-3.5 and GPT-4 outperform the no-context conditions on average. Context both increases the ability of the models to discriminate true from false claims, but additionally, calibrates both models better, predicting true for more cases correctly.

Across all conditions, both models generally fare better in identifying false statements ("pants-fire," "mostly-false," and "false") than true ones ("half-true," "mostly-true," and "true"), mirroring the findings of \cite{hoes2023leveraging} in the no-context settings.
This could be attributed to the inherent complexity of verifying a statement's absolute truth compared to identifying falsehoods.
Lastly, both models have reduced performance in less extreme categories like "half-true" and "mostly-false." These categories are inherently ambiguous and represent statements that have elements of both truth and falsehood, making them more challenging to classify accurately.
The table on the right only considers a claim to be correctly classified when the model predicts the exact category. The extremely low scores show that the models are unable to predict the exact shade of truth that PolitiFact assigned to the statement. 

Both GPT-3.5 and GPT-4 were trained with data up to September of 2021\footnote{\url{https://platform.openai.com/docs/models/gpt-4}}. As the training data of both models is private, data leakage is a significant concern for fact-checks published before that date. To investigate, whether the accuracy of the LLMs differs over time we plot the accuracy of all four conditions (2 models x 2 context conditions) in Fig. \ref{AccPerTime}. 

\setcounter{figure}{4}
\begin{figure}[h]
\includegraphics[width=\textwidth]{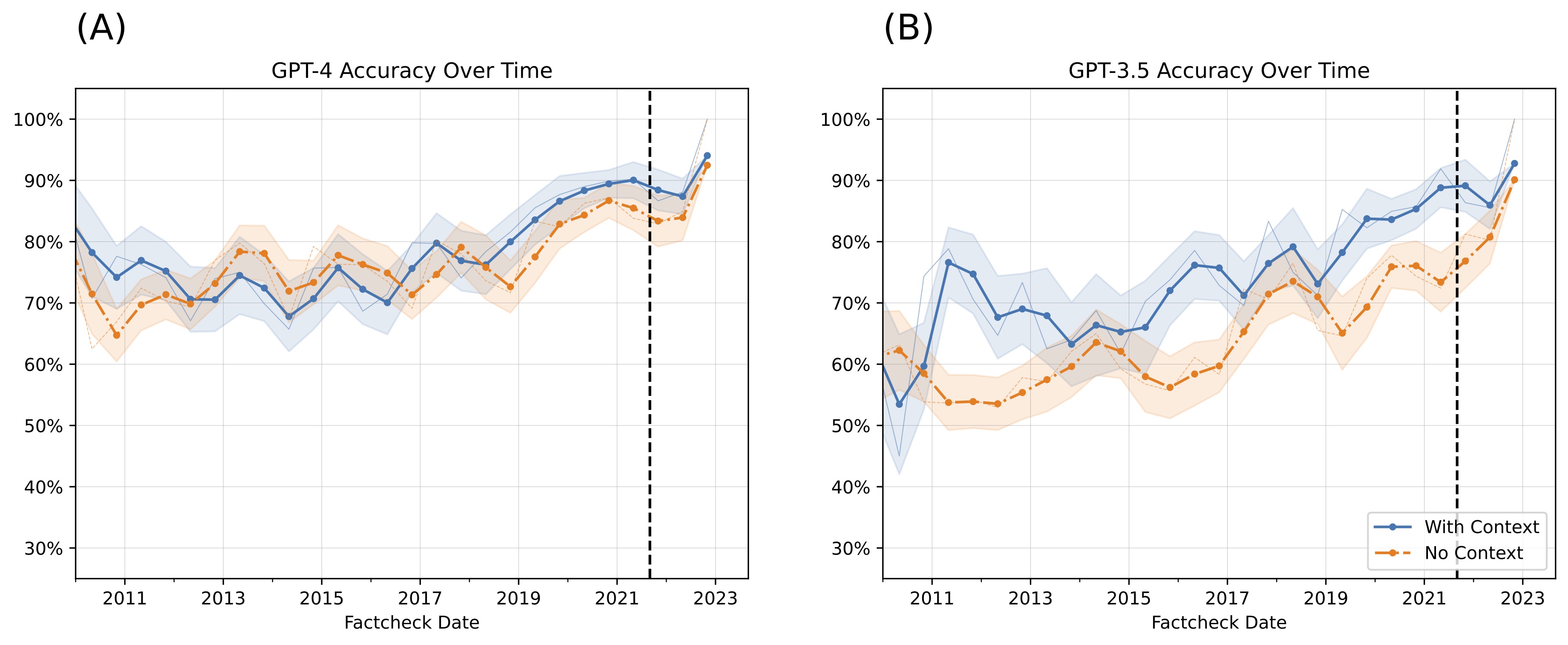}
\caption{Accuracy of GPT-3.5 \& GPT-4 overtime on the PolitiFact dataset.
Yearly rolling average of the accuracy of LLMs over Time. Panel \textbf{(A)} displays the accuracy of GPT-4. Panel \textbf{(B)} shows the accuracy of GPT-3.5. The blue line indicates the context condition, and the orange line indicates the no-context condition. The vertical line represents the training end date of both models according to OpenAI. A faint line is the three-month average. The x-axis represents the date of the claim. The bands represent one standard error.}
\label{AccPerTime}
\end{figure}

The accuracy of all four conditions exhibits an upward trend over time. For the context condition, this improvement can likely be attributed to an increasing availability of relevant information on Google over recent years. More surprisingly, the no-context condition also displays a similar trajectory of improving performance. One potential explanation is that the ground truth labels of some statements in the dataset may have evolved as more facts emerged. We do not observe any sudden decrease in accuracy after the official training cutoff for GPT-3.5 and GPT-4. This suggests that post-training refinements to the models via reinforcement learning from human feedback (RLHF) may introduce new knowledge, particularly for more recent events closer to the RLHF time period.

\subsection{Experiment on the Multilingual Dataset}

\setcounter{table}{1}
\begin{table}[!htbp] \centering 
    \resizebox{\textwidth}{!}{%
\begin{tabular}{@{\extracolsep{5pt}} cccccc} 
\\[-1.8ex]\hline 
\hline \\[-1.8ex] 
Language & Accuracy English & Accuracy Multilingual & F1 English & F1 Multilingual & Number of Samples \\ 
\hline \\[-1.8ex] 
Turkish & 84.19 & 81.50 & 83.59 & 81.91 & $500$ \\ 
Indonesian & 86.68 & 84.59 & 87.52 & 79.59 & $500$ \\ 
French & 74.67 & 81.67 & 75.57 & 72.30 & $166$ \\ 
English & - & 60.98 & - & 68.65 & $500$ \\ 
Thai & 48.21 & 54.34 & 50.41 & 66.45 & $69$ \\ 
Portuguese & 77.22 & 65.56 & 77.02 & 64.28 & $500$ \\ 
Tamil & 70.33 & 59.19 & 67.68 & 51.59 & $500$ \\ 
Spanish & 56.92 & 51.26 & 57.69 & 49.51 & $500$ \\ 
Italian & 45.25 & 50.29 & 48.25 & 47.07 & $427$ \\ 
Chinese & 64.86 & 43.75 & 68.56 & 42.88 & $174$ \\ 
Hebrew & 49.61 & 41.17 & 56.51 & 40.65 & $287$ \\ 
Farsi & 68.28 & 40.54 & 71.96 & 36.26 & $259$ \\ 
Telugu & 59.79 & 26.99 & 67.03 & 27.24 & $500$ \\ 
Azerbaijani & 37.23 & 22.89 & 42.26 & 15.64 & $98$ \\ 
\hline \\[-1.8ex] 
\end{tabular} 
}
\caption{Performance on the multilingual dataset without context.
Accuracy and F1 score for the fact-checking without context on the multilingual dataset. Accuracy English and F1 English are the scores for the case where the models are provided with English translation of the claims while the other scores are computed in the case where the models are provided with the original claims.} 
  \label{Vanilla_Mult} 
\end{table}

Table \ref{Vanilla_Mult} shows the accuracy of GPT-3.5 in evaluating the veracity of fact-checking claims. The columns Accuracy English and Accuracy Multilingual show the percentage of correctly classified claims by language. English refers to the translations of the claims while multilingual refers to the original language. The F1 columns show the F1 score of the models for the Multilingual and English condition. We add the F1 score to the evaluation to account for the fact that the class distribution differs by language.

In all languages, except for Thai, we see an increase in the F1 score of the model when fact-checking the translated claim as compared to the original. This shows that the models that were tested are significantly better at predicting the verdict of a statement when it is presented in English, mirroring prior research indicating that the models struggle to correctly model and classify non-English language text. Similarly, the accuracy was significantly higher for the English language condition.

\setcounter{table}{2}
\begin{table}[!htbp] \centering 
\resizebox{\textwidth}{!}{%
\begin{tabular}{@{\extracolsep{5pt}} cccccc} 
\\[-1.8ex]\hline 
\hline \\[-1.8ex] 
Language & Accuracy English & Accuracy Multilingual & F1 English & F1 Multilingual & Number of Samples \\ 
\hline \\[-1.8ex] 
Portuguese & 87.42 & 89.21 & 75.97 & 80.78 & $500$ \\ 
Indonesian & 87.16 & 89.98 & 77.26 & 73.36 & $500$ \\ 
English & - & 80.16 & - & 71.33 & $500$ \\ 
Telugu & 83.15 & 83.80 & 77.21 & 69.97 & $500$ \\ 
Thai & 77.27 & 50.00 & 55.66 & 66.66 & $69$ \\ 
French & 89.55 & 87.09 & 79.58 & 56.13 & $166$ \\ 
Chinese & 75.81 & 84.78 & 71.86 & 55.35 & $174$ \\ 
Farsi & 77.78 & 59.00 & 69.15 & 48.58 & $259$ \\ 
Turkish & 72.28 & 71.79 & 70.79 & 47.32 & $500$ \\ 
Spanish & 82.68 & 75.18 & 63.01 & 46.30 & $500$ \\ 
Italian & 56.93 & 56.29 & 40.15 & 45.54 & $427$ \\ 
Tamil & 80.56 & 55.22 & 62.95 & 34.86 & $500$ \\ 
Hebrew & 73.61 & 85.71 & 63.81 & 34.42 & $287$ \\ 
Azerbaijani & 62.07 & 44.00 & 43.43 & 33.23 & $97$ \\ 
\hline \\[-1.8ex] 
\end{tabular} 
}
\caption{Performance on the multilingual dataset with context.
Accuracy and F1 score for the fact-checking with context on the multilingual dataset. Accuracy English and F1 English are the scores for the case where the models are provided with English translation of the claims while the other scores are computed in the case where the models are provided with the original claims.} 
\label{Vanilla_Mult_CONTXT} 
\end{table}

\setcounter{figure}{5}
\begin{figure}[!ht]
\centering
\includegraphics[width=0.6\textwidth]{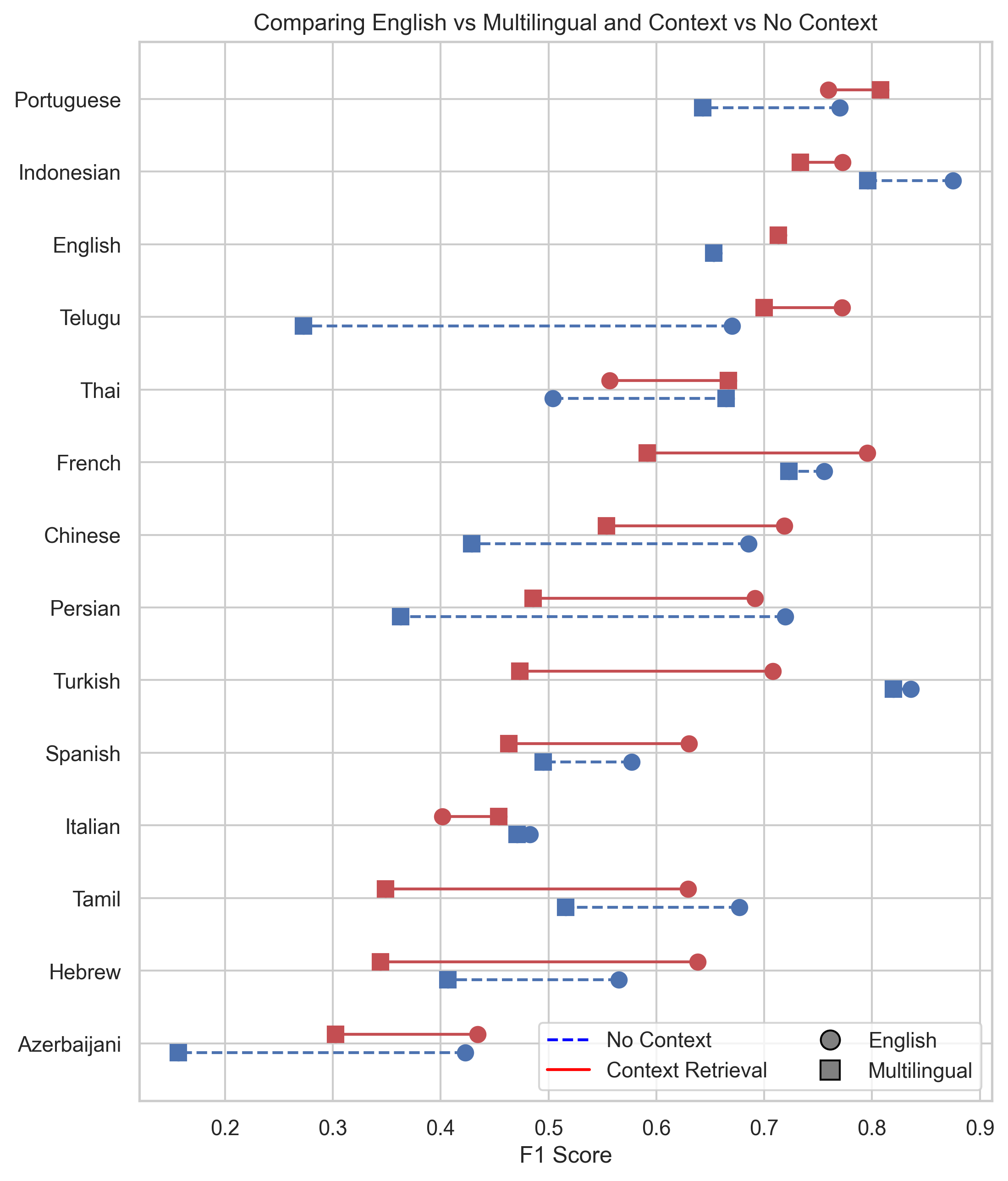}
\caption{Comparison of all conditions on the Data Commons fact-checking dataset. The x-axis displays the F1 score of each language and condition. The y-axis displays each language for which the model was prompted. The blue, dashed line indicates the difference in the performance for the no context condition. The red, solid line shows the difference in performance for the context condition. The circles show the F1 score by language for the translations, the squares show the F1 score for the original scores.}
\label{dumbell}
\end{figure}

Table \ref{Vanilla_Mult_CONTXT} shows the accuracy of GPT-3.5 in evaluating the veracity of fact-checking claims with additional context provided. Similarly to table \ref{Vanilla_Mult} the English language condition is significantly more accurate than the Multilingual condition. In all but three languages (Portugese, Thai, and Italian), translating increased the F1 score of the model. This difference was again significant.

Figure \ref{dumbell} summarizes the findings offered in this section and displays a visualization that compares the F1 scores under the two conditions—'No Context' and 'Context Retrieval'—across various languages. Each language is represented once on the y-axis, with two corresponding horizontal lines plotted at different heights. The blue dashed lines indicate the performance under the 'No Context' condition, while the red solid lines represent the 'Context Retrieval' condition. Each line connects two points, corresponding to the F1 scores achieved when the model is trained on either English-only or Multilingual data. The circle marker displays the F1 score in the English language condition, while the square portrays the efficacy under the multilingual condition.

Similarly, to the PolitiFact experiment, we analyzed the performance of the model over time. Figure \ref{GoogleOverTime} shows the F1 score of all conditions over time. The confidence intervals represent the standard errors and are bootstrapped by sampling with replacement from the dataset and repeatedly calculating the F1 score. As in the PolitiFact experiment, we again see an increase in the performance of the models over time. Mirroring the previous findings, we do not see a decrease in the performance of the models following the official cut-off date for the training of GPT-3.5. 

\setcounter{figure}{6}
\begin{figure}[!ht]
\centering
\includegraphics[width=\textwidth]{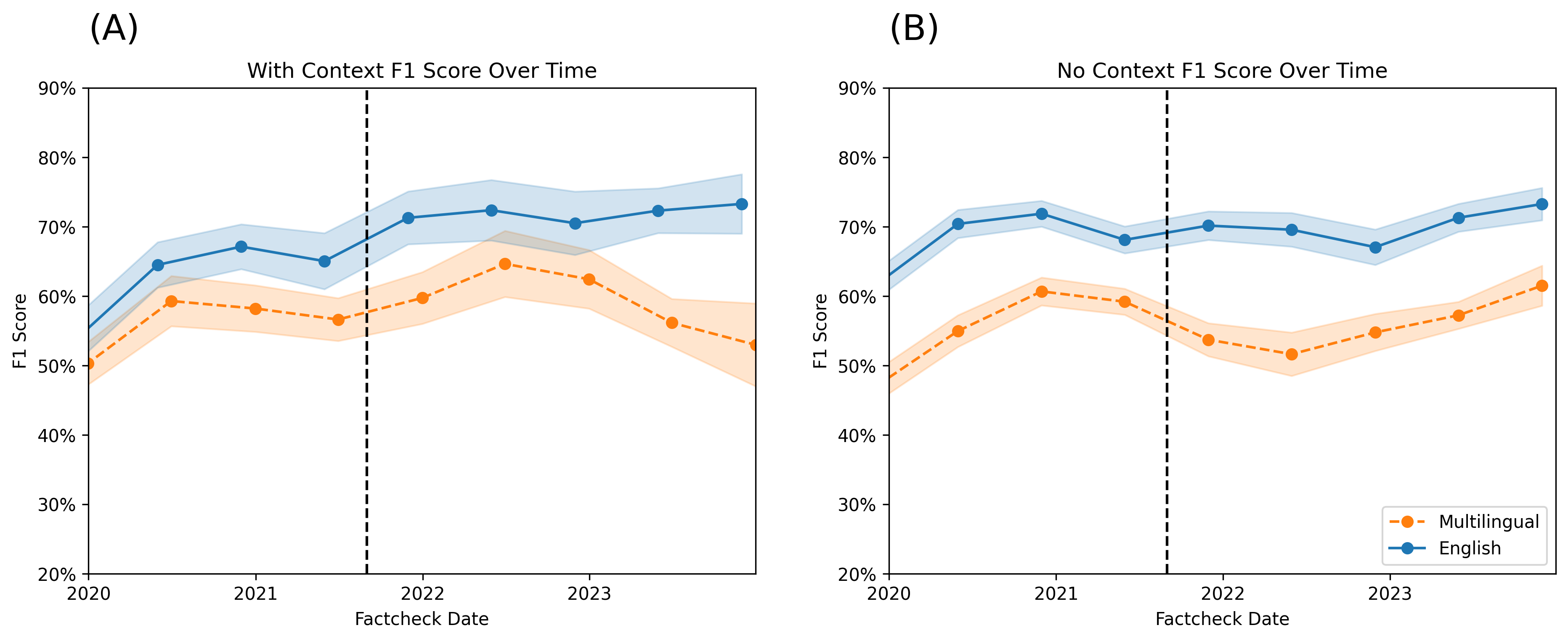}
\caption{F1 score of context \textbf{(A)} and no-context-condition \textbf{(B)} over time. The orange (dashed) line shows the multilingual condition. The blue (solid) line showcases the F1 score for the English condition. The vertical line indicates the official training end date for GPT 3.5.}
\label{GoogleOverTime}
\end{figure}

\section{Conclusion}
In this study, we investigated OpenAIs GPT-3.5 and GPT-4s ability to fact-check claims. We found that GPT-4 outperformed GPT-3.5. Notably, even without contextual information, GPT-3.5 and GPT-4 demonstrate good performance of 63-75\% accuracy on average, which further improves to above 80\% and 89\% for non-ambiguous verdicts when context is incorporated. Another crucial observation is the dependency on the language of the prompt. For non-English fact-checking tasks, translated prompts frequently outperformed the originals, even if the related claims pertained to non-English speaking countries and retrieved predominantly non-English information. The accuracy of these models varied significantly across languages, underscoring the importance of the original language of the claim.

As the training data of the GPT models includes vast amounts of web data, including a filtered version of the Common Crawl Corpus \cite{brown2020language}, there is a significant risk of data leakage for the task of fact-checking. Previous research has shown that misinformation does not exist in isolation, but is repeated \cite{shaar-etal-2020-known} across platforms \cite{micallef2022cross} and languages \cite{kazemi2022matching,quelle2023lost}. As misinformation is repeated and re-occurs, the ability of models to retain previously fact-checked claims can potentially be seen as a benefit rather than a drawback. Nevertheless, this presents a significant risk for the task of fact-checking novel misinformation. We address this in the paper by testing the ability of the models to detect misinformation after the training end date of the models. We found no reduction in performance for fact-checks after the training end date. It seems that incorporating real-time context for novel misinformation enables the models to reason about novel information. In summary, while data-leakage is a significant concern for any application that tests the abilities of LLMs, we argue that for the task of fact-checking it does not seem to degrade performance and might even be beneficial in view of recurring misinformation.

Comparing the performance of the models across languages is inherently difficult, as the fact-checks are not standardised across languages. Fact-checking organisations differ significantly in their choice of which fact-checks to dedicate time to, with some focussing exclusively on local issues and others researching any claims that are viral on social media. Similarly, some fact-checking organisations focus on current claims, while others debunk long-standing misinformation claims. All of these factors influence the ability of a large language model to discern the veracity of a claim. It is therefore possible that fact-checks in low-resource languages perform better than higher resource languages. The most salient point in the analysis of the multilingual misinformation is that the LLMs outperform the multilingual baseline when prompted with the English translation of the fact-checks. The variability in model performance across languages and the improvement of the accuracy when prompted with English language fact-checks indicates that the training regimen, in which the distribution of languages is highly skewed and English is dominant \cite{commoncrawl2023stats}, significantly impacts accuracy. This suggests that the effectiveness of LLMs in fact-checking is not uniformly distributed across languages, likely due to the uneven representation of languages in training data.

Our results suggest that these models cannot completely replace human fact-checkers as being wrong, even if infrequently, may have devastating implications in today's information ecosystem. Therefore, integrating mechanisms allowing for the verification of their verdict and reasoning is paramount.
In particular, they hold potential as tools for content moderation and accelerating human fact-checkers' work.

Looking ahead, it is important to delve deeper into the conditions under which Large Language Models excel or falter. As these models gain responsibilities in various high-stakes domains, it is crucial that their factual reliability is well-understood and that they are deployed judiciously under human supervision. 

While our study concentrated on OpenAI's GPT-3.5 and GPT-4, the rapid evolution of the field means newer, possibly fine-tuned, LLMs are emerging. One salient advantage of our methodology, distinguishing it from others, is the LLM Agents' capability to justify their conclusions. Future research should explore if, by critically examining the reasons and references provided by the LLMs, users can enhance the models' ability to fact-check claims effectively.

\printbibliography  

\end{document}